\documentclass[11pt,a4paper]{article}
\usepackage[hyperref]{naaclhlt2018}
\usepackage{times}
\usepackage{latexsym}
\usepackage{graphicx}
\usepackage{url}
\usepackage{adjustbox}
\usepackage{graphicx}
\usepackage{booktabs}
\usepackage{multirow}
\usepackage{xspace}
\usepackage{float}
\usepackage{xcolor}
\usepackage{rotating}
\usepackage{arydshln}
\usepackage{fancyhdr}
\usepackage[colorinlistoftodos]{todonotes}

\graphicspath{{figures/}}

\definecolor{darkred}{rgb}{0.6,0,0}
\definecolor{darkgreen}{rgb}{0,0.4,0}

\let\oldenumerate\enumerate
\renewcommand{\enumerate}{
	\vspace{-6pt}
	\oldenumerate
	\setlength{\itemsep}{1pt}
	\setlength{\parskip}{0pt}
	\setlength{\parsep}{0pt}
}

\let\oldendenumerate\endenumerate
\renewcommand{\endenumerate}{
	\oldendenumerate
	\vspace*{-6pt}
}

\let\olditemize\itemize
\renewcommand{\itemize}{
	\vspace*{-6pt}
	\olditemize
	\setlength{\itemsep}{1pt}
	\setlength{\parskip}{0pt}
	\setlength{\parsep}{0pt}
}

\let\oldenditemize\enditemize
\renewcommand{\enditemize}{
	\oldenditemize
	\vspace*{-6pt}
}

\renewenvironment{equation}{\vspace{1mm}\begin{oldequation}}{\end{oldequation}\vspace{1mm}}

\makeatletter
\def\adl@drawiv#1#2#3{%
        \hskip.5\tabcolsep
        \xleaders#3{#2.5\@tempdimb #1{1}#2.5\@tempdimb}%
                #2\z@ plus1fil minus1fil\relax
        \hskip.5\tabcolsep}
\newcommand{\cdashlinelr}[1]{%
  \noalign{\vskip\aboverulesep
           \global\let\@dashdrawstore\adl@draw
           \global\let\adl@draw\adl@drawiv}
  \cdashline{#1}
  \noalign{\global\let\adl@draw\@dashdrawstore
           \vskip\belowrulesep}}
\makeatother

\aclfinalcopy %
\def\aclpaperid{644} %

\title{Object Counts! Bringing Explicit Detections Back into Image Captioning}

\author{Josiah Wang, Pranava Madhyastha \and Lucia Specia \\
Department of Computer Science \\
University of Sheffield, UK \\
  {\tt {\{j.k.wang, p.madhyastha, l.specia\}@sheffield.ac.uk} }
}
\date{}

\begin{document}
\maketitle

\thispagestyle{fancy}

\begin{abstract}
The use of explicit object detectors as an intermediate step to image captioning -- which used to constitute an essential stage in early work -- is often bypassed in the currently dominant end-to-end approaches, where the language model is conditioned directly on a mid-level image embedding. We argue that explicit detections provide rich semantic information, and can thus be used as an interpretable representation to better understand why end-to-end image captioning systems work well. We provide an in-depth analysis of end-to-end image captioning by exploring a variety of cues that can be derived from such object detections. Our study reveals that end-to-end image captioning systems rely on matching image representations to generate captions, and that encoding the frequency, size and position of objects are complementary and all play a role in forming a good image representation. It also reveals that different object categories contribute in different ways towards image captioning.
\end{abstract}

\section{Introduction}

Image captioning (IC), or image description generation, is the task of automatically generating a sentential textual description for a given image. %
Early work on IC tackled the task by first running object detectors on the image and then using the resulting explicit detections as input to generate a novel textual description, e.g.\ \cite{KulkarniEtAl:2011,YangEtAl:2011}. With the advent of sequence-to-sequence approaches to IC, e.g.\ \cite{KarpathyFeiFei:2015,VinyalsEtAl:2015}, coupled with the availability of large image description datasets, the performance of IC systems showed marked improvement, at least according to automatic evaluation metrics like Meteor~\cite{DenkowskiLavie:2014} and CIDEr~\cite{VedantamEtAl:2015}. 

The currently dominant neural-based IC systems are often trained end-to-end, using parallel (image, caption) datasets. Such systems are essentially sequential language models conditioned directly on some mid-level image features, such as an image embedding extracted from a pre-trained Convolutional Neural Network (CNN). Thus, they bypass the explicit detection phase of previous methods and instead generate captions \emph{directly} from image features. %
Despite significant progress, it remains unclear why such systems work. 
A major problem with these IC systems is that they are less interpretable than conventional pipelined methods which use explicit detections.

We believe that it is timely to again start exploring the use of explicit object detections for image captioning. Explicit detections offer rich semantic information, which can be used to model the entities in the image as well as their interactions, and can be used to better understand image captioning. %

Recent work \cite{YinOrdonez:2017} showed that conditioning an end-to-end IC model on visual representations that implicitly encode object details yields reasonably good captions. Nevertheless, it is still unclear \emph{why} this works, and \emph{what} aspects of the representation allow for such a good performance. In this paper, we study end-to-end IC in the context of explicit detections (Figure \ref{fig:overview}) by exploring a variety of cues that can be derived from such detections to determine \emph{what} information from such representations helps image captioning, and \emph{why}. To our best knowledge, our work is the first experimental analysis of end-to-end IC frameworks that uses object-level information that is highly interpretable as a tool for understanding such systems. Our main contributions are as follows:

\begin{figure*}
  \begin{center}
  \includegraphics[width=0.8\linewidth]{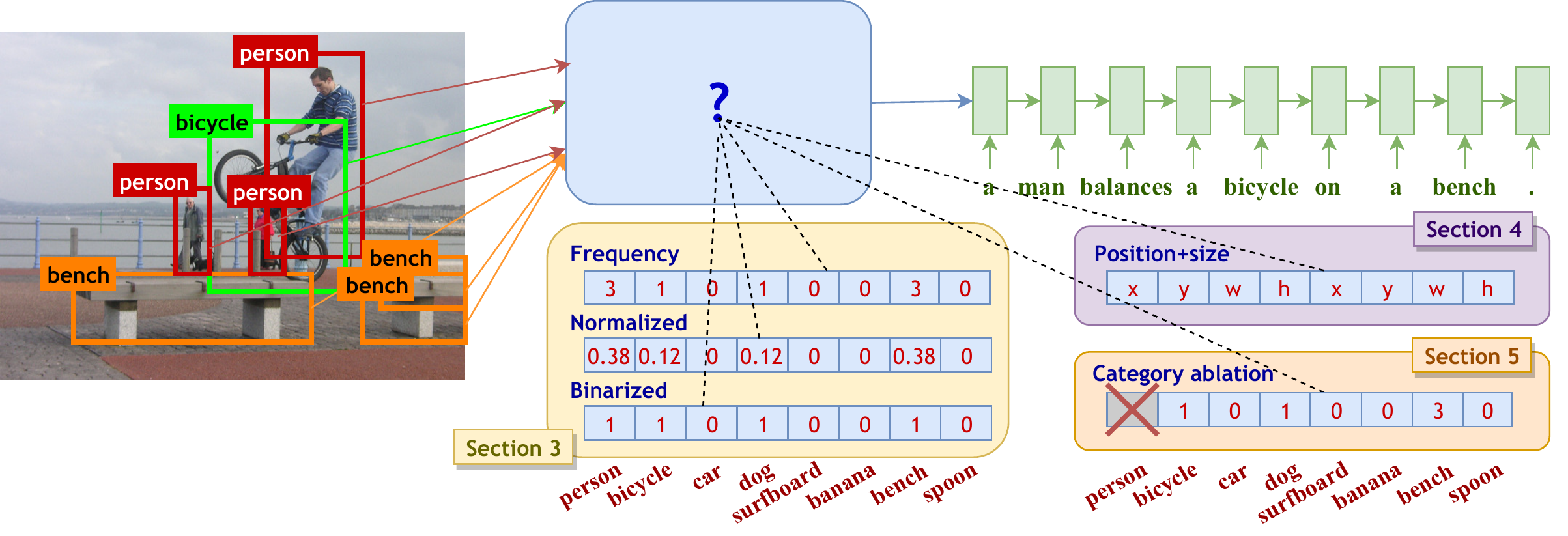}
  \caption{Using explicit detections as an intermediate step towards end-to-end image captioning. The question we investigate is what information can we extract from explicit detections that can be useful for image captioning.
 }
  \label{fig:overview}
  \end{center}
\end{figure*}

\begin{enumerate}
\item We provide an in-depth analysis of the performance of end-to-end IC using a simple, yet effective `bag of objects' representation that is interpretable, %
and generates good captions despite being low-dimensional and highly sparse %
(Section~\ref{sec:boo}).

\item We investigate whether other spatial cues can be used to provide information complementary to frequency counts (Section~\ref{sec:boo}).

\item We study the effect of incorporating different spatial information of individual object instances from explicit detections (Section~\ref{sec:bboo}).

\item We analyze the contribution of the categories in representations for IC by ablating individual categories from them (Section~\ref{sec:categories}).%

\end{enumerate}

Our hypothesis is that there are important components derived from explicit detections that can be used to effectively inform IC. Our study confirms our hypothesis, and that features such as the frequency, size and position of objects all play a role in forming a good image representation to match their corresponding representations in the training set. Our findings also show that different categories contribute differently to IC, and this partly depends on how likely they are to be mentioned in the caption given that they are depicted in the image. The results of our investigation will help further work towards more interpretable  image captioning. %

\section{Related work}
\label{sec:relatedwork}

Early work on IC apply object detectors explicitly on an image as a first step to identify entities present in the image, and then use these detected objects as input to an image caption generator.  
The caption generator typically first performs content selection (selecting a subset of objects to be described) and generates an intermediate representation (e.g.\ semantic tuples or abstract trees), and then performs surface realization using rules, templates, $n$-grams or a maximum entropy language model. The main body of work uses object detectors for 20 pre-specified PASCAL VOC (Visual Object Classes)~\cite{EveringhamEtAl:2015}  \cite{YangEtAl:2011,KulkarniEtAl:2011,LiEtAl:2011,MitchellEtAl:2012}, %
builds a detector inferred from captions \cite{FangEtAl:2015}, or assumes gold standard annotations are available \cite{ElliottKeller:2013,YatskarEtAl:2014}.

Currently, deep learning end-to-end approaches dominate IC work~\cite{DonahueEtAl:2015,KarpathyFeiFei:2015,%
VinyalsEtAl:2015}. Such approaches do not use an explicit detection step, but instead use a `global' image embedding as input (generally %
a CNN) and learn a language model %
(generally an LSTM) conditioned on this input. Thus, they are trained to learn image caption generation directly from a parallel image--caption dataset. %
The advantage is that no firm decisions need to be made about object categories. However, such approaches are hard to interpret and are dataset dependent~\cite{VinyalsEtAl:2017}. %

Some recent work use  %
object-level semantics for end-to-end IC~\cite{GanEtAl:2017:Semantic,WuEtAl:2016,YouEtAl:2016}. %
Such systems represent images as predictions of semantic concepts occurring in the image. These predictions, however, are at a global, \emph{image} level (``does this image contain a chair?''), rather than at object \emph{instance} level (``there is a big chair at position $x$''). In addition, most previous work regard surface-level terms extracted directly from captions as `objects', while we use off-the-shelf predefined object categories which have a looser connection between the image and the caption (e.g.\ objects can be described in captions using different terms, depicted objects might not be mentioned in captions, and captions might mention  objects that are not depicted).

\newcite{YinOrdonez:2017} propose conditioning an end-to-end IC model on information derived from explicit detections. They implicitly encode the category label, position and size of %
object instances as an `object-layout' LSTM and condition the language model on the final hidden state of this LSTM, and produce reasonably good image captions based only on those cues, without the direct use of images.  
Our work is different in that we feed information from explicit object detections \emph{directly} to the language model in contrast to an object-layout LSTM which abstracts away such information, thereby retaining the \emph{interpretability} of the input image representation. This gives us more control over the image representation which is simply encoded as a bag of categorical variables. 

There is also recent work applying attention-based models~\cite{XuEtAl:2015} on explicit object proposals ~\cite{AndersonEtAl:2017,LiEtAl:2017}, which may capture object-level information from the attention mechanism. However, attention-based models require object information in the form of vectors, whereas our models use information of objects as categorical variables which allow for easy manipulation but are not compatible with the standard attention-based models. 
The model that we use, \emph{under similar conditions} (i.e.\ under similar parametric settings), is comparable to the state-of-the-art models. %

\section{Bag of objects}
\label{sec:boo}

We base our experiments on the MS COCO dataset~\cite{LinEtAl:2014}. From our %
preliminary experiments, we found that a simple \emph{bag of object categories} used as an image representation for end-to-end IC led to good scores according to automatic metrics, comparable to and perhaps even higher than those using CNN embeddings. This is surprising given that this bag of objects vector is \emph{low-dimensional} (each element represents the frequency of one of 80 COCO categories) and \emph{sparse} (mainly zeros, as only a few object categories tend to occur in a given image). In simple terms, it appears that the IC model can generate a reasonable caption by merely knowing what is in the image, e.g. that there are three \textit{person}s, three \textit{benches}
and a \textit{bicycle} in Figure~\ref{fig:overview}. 

This observation raises the following questions. What is it in this simple bag of objects representation that contributes to the surprisingly high performance on IC? Does it lie in the frequency counts? Or the choice of categories themselves? %
It is also worth noting that the image captions in COCO were crowd-sourced \emph{independent} of the COCO object annotations, i.e.\ image captions were written based only on the image, without object-level annotations. The words used in the captions thus do not correspond directly to the 80 COCO categories (e.g.\ a \textit{cup} may not be mentioned in a description even though it is present in the image, and vice versa, i.e. objects described in the caption may not correspond to any of the %
categories).

In order to shed some light into what makes bag of object categories representations work so well for IC, we first investigate whether the \emph{frequency counts} is the main contributor. We then proceed to studying what else can be exploited from explicit object detections to improve on the bag of objects model, for example the size of object instances. We also perform an analysis on these representations to gain more insights into why the bag of objects model performs well.

\subsection{Image captioning model}
\label{sec:model}

Our implementation is based on the end-to-end approach of~\newcite{KarpathyFeiFei:2015}. We use an LSTM~\cite{HochreiterSchmidhuber:1997} language model as described in \newcite{ZarembaEtAl:2014}. To condition the image information, we first perform a linear projection of the image representation followed by a non-linearity:
\begin{equation}
x = \sigma{(W{\cdot}I_m)}
\label{affine}
\end{equation}
\noindent{where $I_m \in \mathcal{R}^d$ is the $d$-dimensional initial image representation, $W \in \mathcal{R}^{n{\times}d}$ is the linear transformation matrix, $\sigma$ is the non-linearity. We use  Exponential Linear Units~\cite{ClevertEtAl:2015} as the non-linear activation in all our experiments. %
We initialize the LSTM-based caption generator with the projected image representation, $x$.} %

\paragraph{Training and inference.}

The caption generator is trained to generate sentences conditioned on %
$x$. We train the model by minimizing the cross-entropy, i.e. the sentence-level loss corresponds to the sum of the negative log likelihood of the correct word at each time step: 
\begin{equation}
\Pr(S{\mid}x;\theta) = \sum_t{\log(\Pr(w_t|w_{t-1}..w_0;x))}
\label{likelihood}
\end{equation}
\noindent{where $\Pr{(S{\mid}x;\theta)}$ is the sentence-level loss conditioned on the image feature $x$ and $\Pr(w_t)$ is the probability of the word at time step $t$. 
This is trained with standard teacher forcing as described in \newcite{SutskeverEtAl:2014}, where the correct word information is fed to the next state in the LSTM. 

Inference is usually performed using approximate techniques like beam search and sampling  methods. %
As we are mainly interested in studying %
different image representations, we focus on the language output that the models can most confidently produce. In order to isolate any other variables from the experiments, we generate captions using a greedy $\arg\max$ approach. %
We use a 2-layer LSTM with $128$-dimensional word embeddings and $256$-dimensional hidden dimensions. As training vocabulary we retain only words that appear at least twice. We provide details about hyperparameters and tuning in Appendix~\ref{appendix:hyperparameter}.

\subsection{Visual representations}
\label{sec:representations}

The first part of our experiments studies the role of \emph{frequency counts} of the 80-dimensional bag of objects representation. We explore the effects of using the following variants of the bag of objects representation:
(i)~\textbf{Frequency:} The number of instances per category;
(ii) \textbf{Normalized:} The frequency counts normalized such that the vector sums to 1. This represents the \emph{proportion} of object occurrences in the image;
(iii) \textbf{Binarized:} An object category's entry is set to 1 if at least one instance of the category occurs, and 0 otherwise. 

\newcite{BergEtAl:2012} explore various factors that dictate what objects are mentioned in image descriptions, and found that object size and its position relative to the image centre are important. Inspired by these findings, we explore  %
alternative representations based on these cues:
(i) \textbf{Object size:} The area of the region %
provided by COCO, normalized by %
image size; we encode the largest object if multiple objects occur for the same category %
(max pooling). %
(ii) \textbf{Object distance:} The Euclidean distance from the object bounding box centre to the image centre, normalized by %
image size; we encode the object closest to the centre if multiple instances occur %
(min pooling). 
We also explore \textbf{concatenating} these 
features to study their complementarity.

Finally, we study the effects of \textbf{removing information} from the bag of objects representation. More specifically, we compare the results of retaining only a certain number of object instances in the \emph{frequency}-based bag of objects representation, rather than representing an image with all objects present. We experiment with retaining only the frequency counts for one object category and $25\%$, $50\%$, and $75\%$ of object categories; the remaining entries in the vector are set to zero. The object categories to be retained are selected, per image: (i) randomly; (ii) by the $N\%$ most frequent categories of the image; (iii) by the $N\%$ largest categories of the image; (iv) by the $N\%$ categories closest to the centre of the image.

We performed these evaluations based on (i) ground truth COCO annotations and (ii) the output of an off-the-shelf object detector~\cite{RedmonFarhadi:2017} trained on 80 COCO categories. With ground truth annotations we can isolate issues stemming from incorrect detections.

\subsection{Experiments}

\begin{table}[t]
\begin{center}
{ \small
\begin{tabular}{lcc}
\toprule
Representation & \textbf{GT} & \textbf{Detect} \\
\midrule
\textbf{CNN (ResNet-152 POOL5)} & - & 0.749 \\
\midrule
\textbf{Frequency} & 0.807 & 0.752 \\
\textbf{Normalized} & 0.762 & 0.703 \\
\textbf{Binarized} & 0.751 & 0.703 \\
\midrule
\textbf{Object min distance} & 0.759 & 0.691\\
\textbf{Object max size} & 0.793 & 0.725 \\
\midrule
\textbf{Obj max size + Obj min distance} & 0.799 & 0.743 \\
\textbf{Frequency + Obj min distance} & 0.830 & 0.769\\
\textbf{Frequency + Obj max size} & 0.836 & 0.769\\
\textbf{All three features} & 0.849 & 0.743\\
\bottomrule
\end{tabular}
}
\end{center}
\caption{CIDEr scores for image captioning using bag of objects variants as visual representations. We compare the results of using ground truth annotations (\textbf{GT}) and the output of a detector (\textbf{Detect}). As comparison we also provide, in the first row, the results of using a ResNet-152 POOL5 CNN image embedding with our implementation of an end-to-end IC system.}
\label{tbl:boo-results}
\end{table}

We train our models on the full COCO training set, and use the standard, publicly available splits\footnote{\url{http://cs.stanford.edu/people/karpathy/deepimagesent}} of the validation set as in previous work~\cite{KarpathyFeiFei:2015} for validation and testing (5,000 images each). We use CIDEr~\cite{VedantamEtAl:2015} %
-- the official metric for COCO -- as our evaluation metrics for all experiments. %
For completeness, we present scores for other common IC metrics in Appendix~\ref{appendix:fullresults}.

\begin{table}[!t]
\begin{center}
{\small
\begin{tabular}{lccc}
\toprule
Feature vs. Pooling & \textbf{Min} & \textbf{Max} & \textbf{Mean} \\
\midrule
\textbf{Obj. Size} & 0.748 & 0.793 & 0.789 \\
\textbf{Obj. Distance} & 0.759 & 0.768 & 0.740\\
\bottomrule
\end{tabular}
}
\end{center}
\caption{CIDEr scores for captioning comparing the use of min, max or average pooling of either object size or distance features, using ground truth annotations.}
\label{tbl:boominmax-results}
\end{table}

Table~\ref{tbl:boo-results} shows the CIDEr scores of IC systems using variants of the bag of objects representation, for both ground truth annotations and the output of an object detector. Compared to a pure CNN embedding (ResNet-152 POOL5), our object-based representations show higher (for ground truth annotations) or comparable CIDEr scores (for detectors).
Our first observation is that frequency \textbf{counts} are essential to IC. Using normalized counts as a representation gives poorer results, which intuitively makes sense: An image with 20 cars and 10 people is significantly different from an image with two cars and one person. Using binarized counts (presence or absence) brings the score further down. This is to be expected: An image with one person is very different from one with 10 people. 

Using spatial information (size or distance) also proved useful. Encoding the object size in place of frequency gave reasonably better results over using object distance from the image centre. We can conclude that the size and centrality of objects are important factors for captioning, with object size being more informative than position.

We also experimented with different methods for aggregating multiple instances of the same category, in addition to choosing the biggest instance and the instance closest to the image centre. For example, choosing the \emph{smallest} instance (min pooling) or the instance \emph{furthest} away from the image centre (max pooling), or just averaging them (mean pooling). Table~\ref{tbl:boominmax-results} shows the results. %
For object size, the findings are as expected: Smaller object instances are less important for IC, although averaging them works comparably well. Surprisingly, in the case of distance, using the object \emph{furthest} from the image centre actually gave slightly better results than the one closest. 
Further inspection revealed that %
aggregating instances is not effective in some cases. We found that the positional information (and interaction with other objects) captured by the object further away may sometimes represent the semantics of the image better than the object in the centre of the image. For example, in Figure~\ref{fig:pooling_example}, encoding only the position of the person in the middle will result in the representation being similar to other images with only one person in the centre of the image (and also on a kitchen counter).
Representing the person as the one furthest from the image will result in some inference (from training data) that there could be more than one person in the image sitting \emph{around} the kitchen counter rather than a single person standing \emph{at} the kitchen counter.

\begin{figure}[!t]
  \begin{center}
  { \small
  \includegraphics[width=0.58\linewidth]{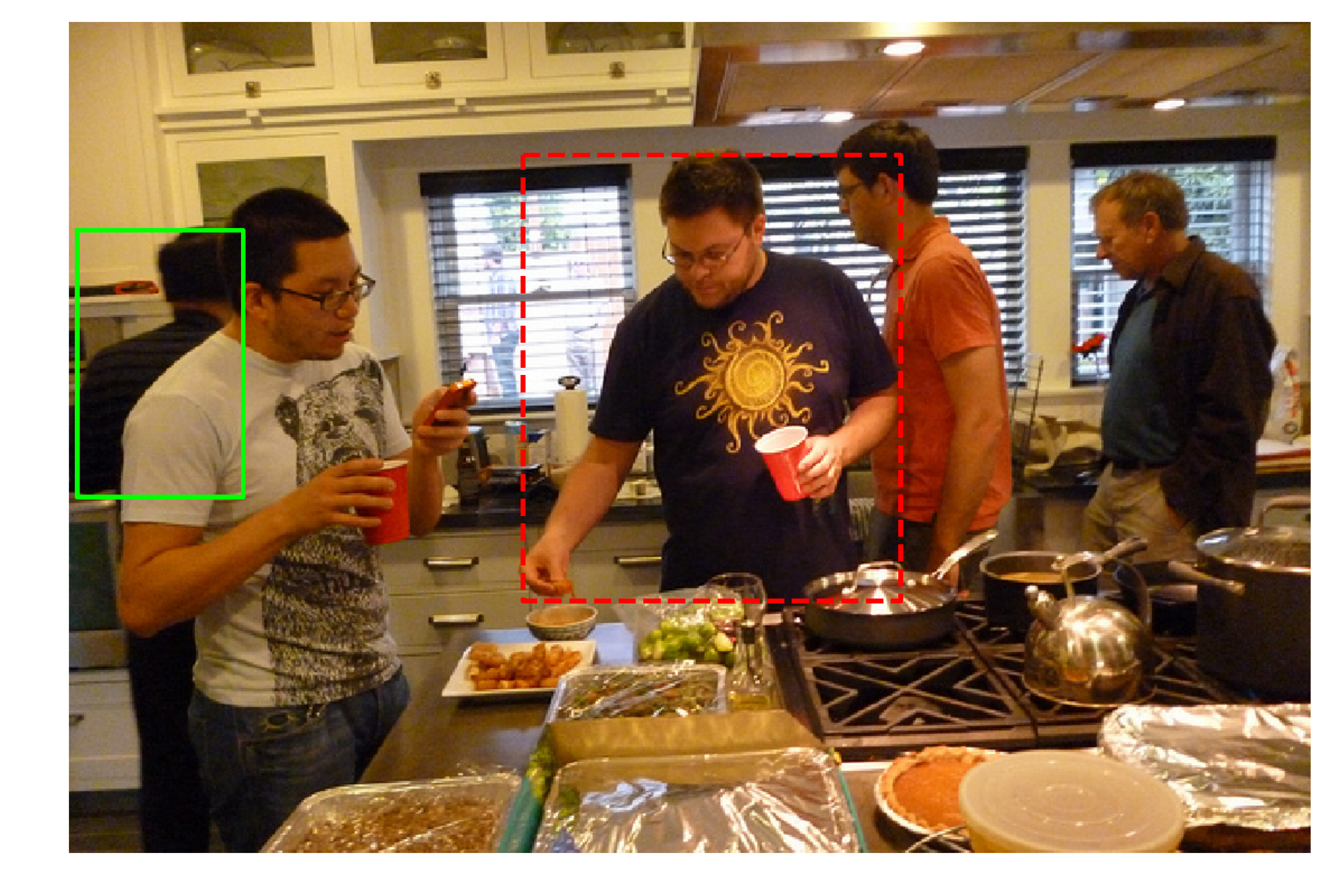} 

  \begin{minipage}{1.0\linewidth}
    \textcolor{darkred}{Obj. min distance:}\\
    $\bullet$ a man in a kitchen preparing food in a kitchen .\\
    \textcolor{darkgreen}{Obj. max distance:}\\ 
    $\bullet$ a group of people standing around a kitchen counter .\\
   \end{minipage}
  }
  \caption{Example where encoding the distance of the object furthest away (solid \textcolor{darkgreen}{green}) is better than that of the one closest to the image centre (dashed \textcolor{darkred}{red}). The IC model assumes that only one person is in the middle in the former case, and infers that many people may be gathered around a table in the latter.}
  \label{fig:pooling_example}
  \end{center}
\end{figure}

The combination of results (bottom row of Table~\ref{tbl:boo-results})  shows that the three features (frequency, object max size and min distance) are complementary, and that combining any pair gives better CIDEr scores than each alone. The combination of all three features produces the best results. These results are interesting, as adding spatial information of even just one object per category can produce a better score. This has, to our knowledge, not been previously demonstrated.
The performance of using an explicit detector rather than ground truth annotations is poorer, as expected from noisy detections. However, the overall trend generally remains similar, except for the combination of all three features which gave poorer scores. %

Finally, Figure~\ref{fig:boothreshold-results} shows the results of partially removing or masking the information captured by the bag of object representation (frequency). As expected, IC performance degrades when less than $75\%$ of information is retained. The performance of the system where the representation is reduced using frequency information suffers the most (even worst than removing categories randomly), suggesting that frequency does not correspond to an object category's importance, i.e.\ just because there is only one \textit{person} in the image does not mean that it is less important than the ten \textit{car}s depicted. On the other hand, object size correlates with object importance in IC, i.e.\ larger objects are more important than smaller objects for IC: The performance does not degrade as much as removing categories by their frequency in the image.

\begin{figure}[!t]
\begin{center}
\includegraphics[width=0.90\linewidth]{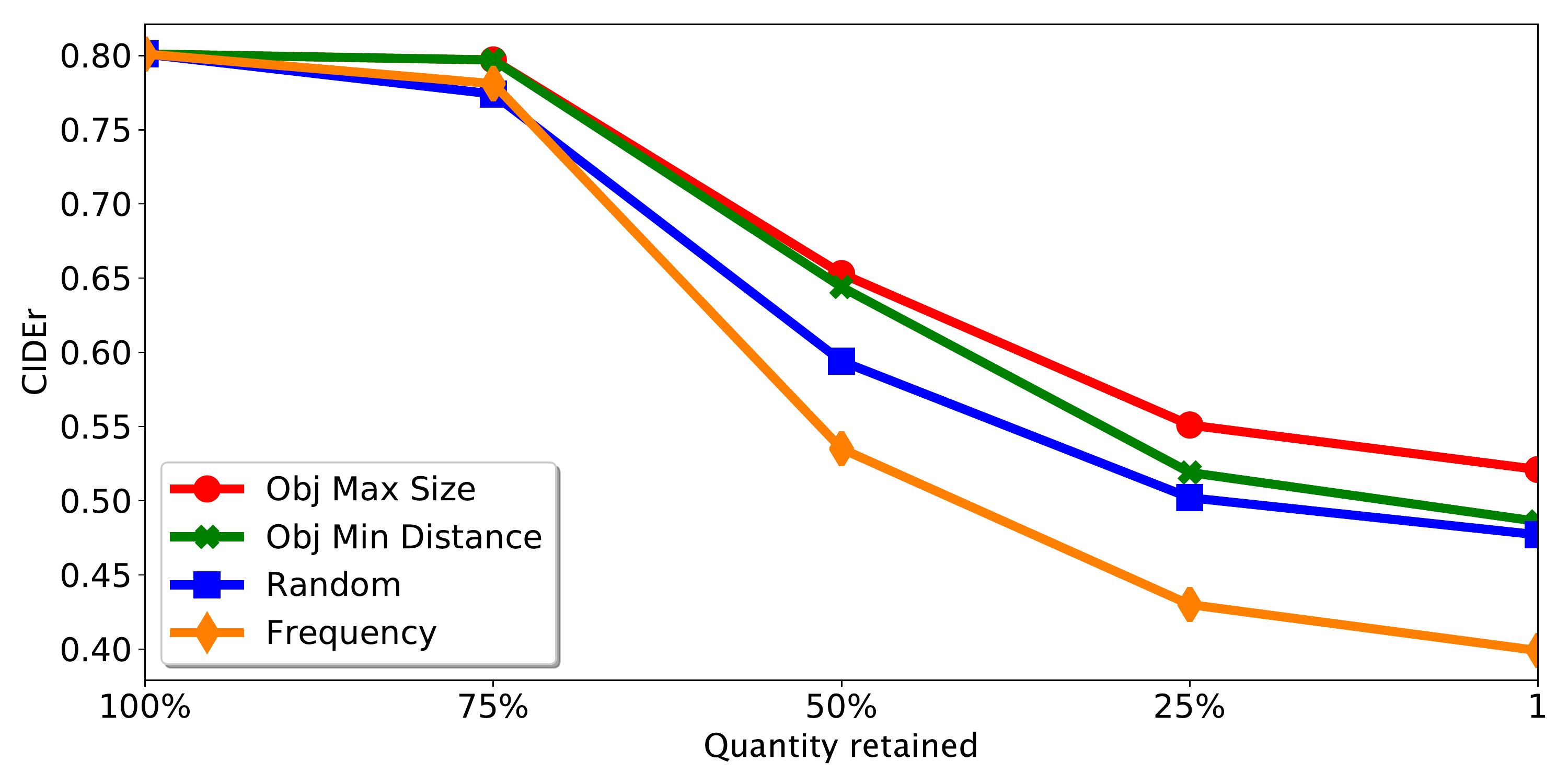}
\end{center}
\vspace{-8pt}
\caption{Change in CIDEr scores for image captioning by reducing the number of (ground truth) object instances in the image representation, based on different heuristics.} 
\label{fig:boothreshold-results}
\end{figure}

\subsection{Analysis}
\label{sec:analysis}

We hypothesize that the bag of objects representation performs well because it serves as a good representation for the dataset and allows for better image matching. One observation is that the category distribution between the training and test sets are very similar (Figure~\ref{fig:object_distribution})%
, thus increasing the chance of the bag of objects representation producing a close match to one in the training set. 
From this observation, we posit that end-to-end IC models leverage COCO being repetitive to find similar matches for a test image to a combination of images in the training set. Further investigation on the category distribution (e.g.\ by splitting the dataset such that the test set contains unseen categories) is left for future work. %

\paragraph{$k$-Nearest neighbour analysis.}
We further investigate our claim that end-to-end IC systems essentially perform complex image matching against the training set with the following experiment. The idea is that if the IC model performs some form of image matching and text retrieval from the training set, then the nearest neighbour (from training) of a test image should have a caption similar to the one generated by the model. However, 
the model does not always perform text retrieval as the LSTM is known to sometimes generate novel captions, %
possibly by aggregating or `averaging' the captions of similar images and performing some factorization. We first generate captions for every training image using the bag of objects model (with ground truth frequency counts). We then compute the $k$-nearest training images for each given test image using both the bag of objects representation and its projection (Eq.~\ref{affine}). Finally, we compute the similarity score between the generated caption of the test image against all $k$-nearest captions. The similarity score measures how well a generated caption matches its nearest neighbour's captions. We expect the score to be high if 
the IC system generates an image similar to something `summarized' from the training set.

\begin{figure}[!t]
  \begin{center}
  \includegraphics[width=1.0\linewidth]{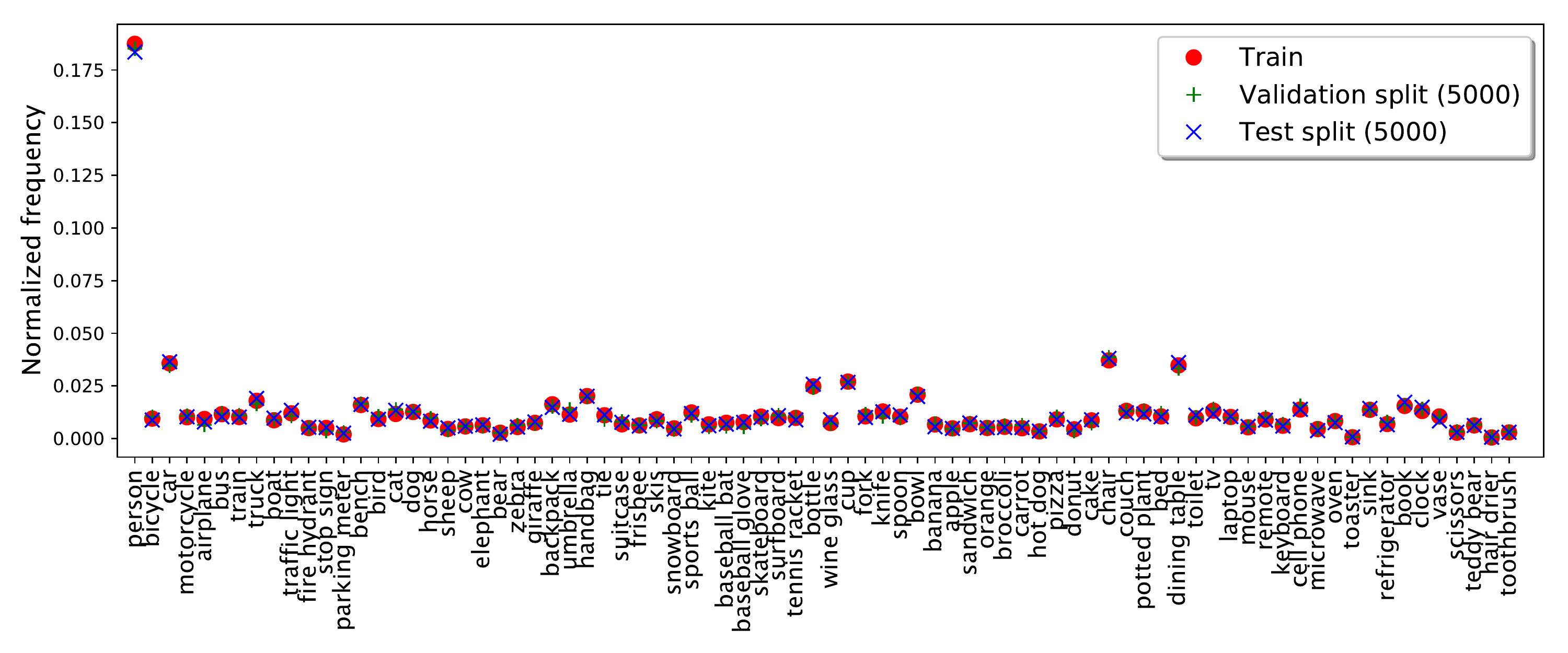}
  \end{center}
  \vspace{-12pt}
  \caption{Object category distributions for COCO train, validation and test splits:  normalized document frequency of each category. The distribution between the training and test sets are almost identical. A higher resolution version can be found in Appendix~\ref{appendix:fullresults}.}
  \label{fig:object_distribution}
\end{figure}

\begin{table}[t]
\begin{center}
{\small
\begin{tabular}{l c c c c}
\toprule
Type & BLEU & Meteor & CIDEr & SPICE\\
\midrule
Freq. & 0.868 & 0.591 & 6.956 & 0.737\\
Proj. & 0.912 & 0.634 & 7.651 & 0.799\\
\midrule
Exact ($2301$) & 1.000 & 1.000 & 10.000 & 1.000\\
\midrule
Freq. ($\neg$ Exact) & 0.757 & 0.498 & 4.337 & 0.512 \\
Proj. ($\neg$ Exact) & 0.837 & 0.560 & 5.638 & 0.628 \\
\bottomrule
\end{tabular}
}
\end{center}
\caption{$k$-Nearest Neighbour ($k$=5) trial on the ground truth bag of objects (\textbf{Freq.}) and the projected bag of objects (\textbf{Proj.}) representations. The references are captions of $5$-nearest images in each space. \emph{Exact} represents a subset of 2301 samples where all the 5 neighbours have $0$ distance (replicas) and $\neg$ represents nearest neighbours that are not replicas of the test image.}
\label{tbl:knn}
\end{table}

As reported in Table~\ref{tbl:knn},  overall the captions seem to closely match the captions of $5$ nearest training images. Further analysis showed that 2301 out of 5000 captions had nearest images at a zero distance, i.e., the same exact representation was seen at least $5$ times in training %
(note that CIDEr gives a score of 10 only if the test caption and \emph{all} references are the same). We found that among the non-exact image matches, the projected image representation  better captures candidates in the training set than bag of objects. Figure~\ref{fig:knn-example} %
shows the five nearest neighbours of an example non-exact match and their generated captions in the projection space. Note that the nearest neighbours are an approximation since we do not know the exact distance metric derived from the LSTM. We observe that the captions for unseen representations seem to be interpolated from multiple neighbouring points in the projection space, but further work is needed to analyze the hidden representations of the LSTM to understand the language model and to give firmer conclusions.%

\begin{figure}[!t]
\begin{center}
{ \tiny
\begin{tabular}{p{0.000001cm} p{1.5cm} p{5cm}}
\toprule
\multirow{4}{*}{\begin{sideways}test\end{sideways}} & \multirow{4}{*}{\includegraphics[height=0.8cm]{COCO_val2014_000000242946.jpg}} & \textit{person~(5), cup~(8), spoon~(1), bowl~(8), carrot~(10), chair~(6), dining table~(3)} \\
& & \vspace{-2pt} $\Rightarrow$ a group of people sitting around a table with food . \\
&& \\
\midrule
\multirow{4}{*}{1} &  \multirow{4}{*}{\includegraphics[height=0.8cm]{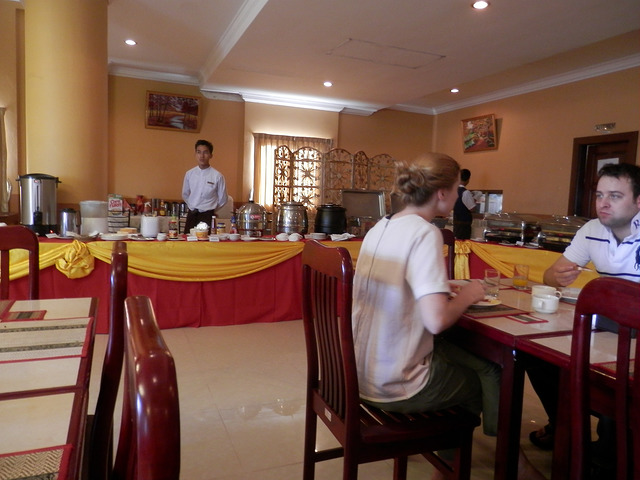}} & \textit{person~(4), cup~(4), spoon~(1), bowl~(5), chair~(6), dining table~(4)} \\
& & \vspace{-2pt} $\Rightarrow$ a woman sitting at a table with a plate of food . \\
& & \\
\multirow{4}{*}{2} & \multirow{4}{*}{\includegraphics[height=0.8cm]{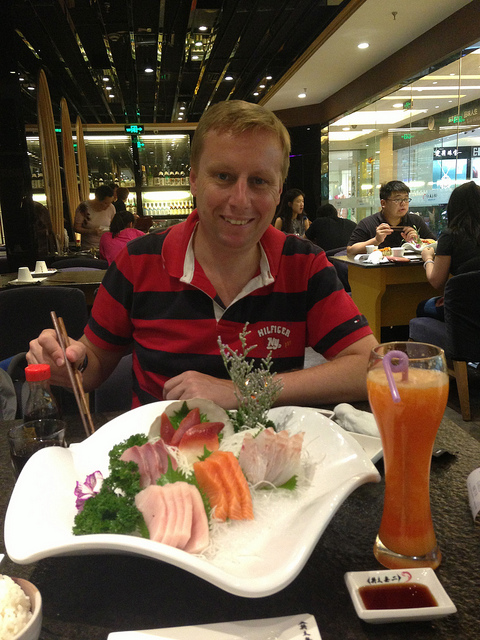}} & \textit{person~(9), bottle~(1), cup~(6), bowl~(4), broccoli~(2), chair~(5), dining table~(3)} \\
& & \vspace{-2pt} $\Rightarrow$ group of people sitting at a table eating food . \\
& & \\
\multirow{4}{*}{3} & \multirow{4}{*}{\includegraphics[height=0.8cm]{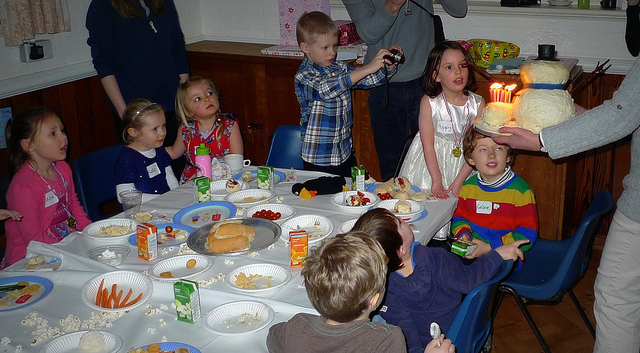}} & \textit{person~(11), cup~(2), bowl~(4), carrot~(6), cake~(1), chair~(4), dining table~(1)} \\
& & \vspace{-2pt} $\Rightarrow$ a group of people sitting around a table with a cake . \\
& & \\
\multirow{4}{*}{4} & \multirow{4}{*}{\includegraphics[height=0.8cm]{COCO_train2014_000000093992.jpg}} & \textit{cup~(1), spoon~(1), bowl~(9), carrot~(10), chair~(3), potted plant~(1), dining table~(1), vase~(1)} \\
& & \vspace{-2pt} $\Rightarrow$ a table with a variety of food on it . \\
& & \\
\multirow{4}{*}{5} & \multirow{4}{*}{\includegraphics[height=0.8cm]{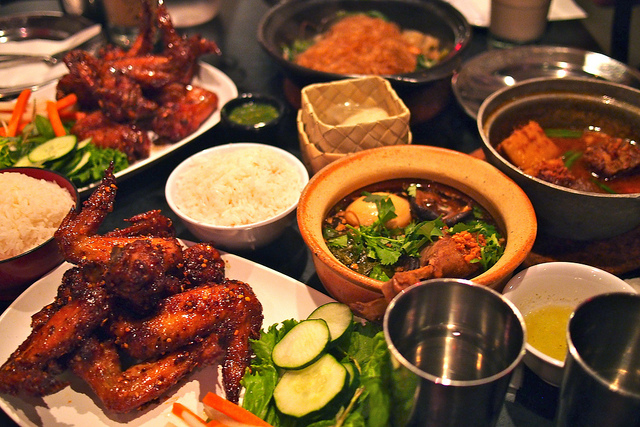}} & \textit{cup~(4), bowl~(7), carrot~(6), dining table~(1)} \\
& & \vspace{-2pt} $\Rightarrow$ a table with bowls of food and vegetables . \\
& & \\
\noalign{\smallskip}
\bottomrule
\end{tabular}
}
\end{center}
\caption{Five nearest neighbours from the training set in the projected space, for an example test image's (top row) original bag of objects representation that does not have an exact match in the training. For each image, we show the ground truth categories (and frequencies in parenthesis) and the generated caption. More examples can be found in Appendix~\ref{appendix:knn-moreexample}.}
\label{fig:knn-example}
\end{figure}

\section{Spatial information on instances}
\label{sec:bboo}

Here we further explore the effect of incorporating spatial information of object detections for IC. More specifically, we enrich the representations by encoding positional and size information for more object instances, rather than restricting the encoding to only one instance per category which makes the %
representation less informative.

\subsection{Spatial representation}

We explore encoding \textbf{object instances and their spatial properties} as a fixed-size vector. In contrast to Section~\ref{sec:boo}, we propose handling multiple instances of the same category by encoding spatial properties of \emph{individual} instances rather than aggregating them as a single value. Each instance is represented as a tuple ($x$, $y$, $w$, $h$, $a$), where $x$ and $y$ are the coordinates of the centre of the bounding box and are normalized to the image width and height respectively, $w$ and $h$ are the width and height of the bounding box respectively, and $a$ is the area covered by the object segment and normalized to the image size. Note that $w \times h \geq a$ (box encloses the segment). We assume that there are maximum 10 instances per vector, and instances of the same category are ordered by $a$ (largest instance first). We encode each of the 80 categories as separate sets. Non-existent objects are represented with zeros. The dimension of the final vector is $4000$ ($80 \times 10 \times 5$). We also perform a \textbf{feature ablation} experiment to isolate the contribution of different spatial components. %

\begin{figure*}[!t]
\begin{center}
{\scriptsize
\begin{minipage}{0.2\textwidth}
\includegraphics[width=0.9\textwidth]{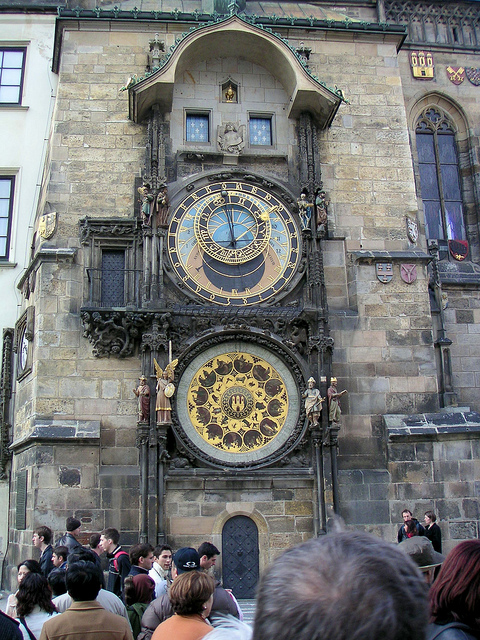}
Image ID: 378657
\end{minipage}
\begin{minipage}{0.7\textwidth}
\begin{tabular}{p{0.25\linewidth}p{0.75\linewidth}}
\toprule
\textbf{Objects in the image}  & \emph{person}, \emph{clock} \\
\midrule
\textbf{Representation} & \textbf{Caption}\\
\midrule
Frequency & a large clock tower with a large clock on it . \\
Object min distance & a clock tower with a large clock on it 's face . \\
Object max size & a man standing in front of a clock tower . \\
All three features &  a clock tower with people standing in the middle of the water . \\
$(x,y)$ &  a large clock tower with a clock on the front . \\
$(w,h)$ & a clock on a pole in front of a building\\
$(a)$ &  a large clock tower with people walking around it\\
$(x,y,w,h,a)$ & a group of people standing around a clock tower .\\ 
\cdashlinelr{1-2}
CNN (ResNet-$152$) & a large building with a clock tower in the middle of it .\\
\textit{person} removed & a clock tower with a weather vane on top of it .\\
\bottomrule
\end{tabular}
\end{minipage}
}
\end{center}
\caption{Example captions with different models. The models with explicit object detection and additional spatial information ($(x,y,w,h,a)$) are more precise in most cases. The output of a standard ResNet-152 POOL5 is also shown, as well as that of the model where the most salient category -- \textit{person} -- is removed from the feature vector. More example outputs are available in Appendix~\ref{appendix:example-output}.}
\label{fig:examples}
\end{figure*}

\subsection{Experiments}

All experiments in this subsection use ground truth annotations -- we expect the results of using an object detector to be slightly worse but in most cases follow a similar trend, as shown in the previous section. %
Table~\ref{tbl:bboo-results} shows the CIDEr scores using the same setup as Section~\ref{sec:boo}, but using representations with spatial information about individual object instances. Encoding spatial information led to substantially better performance over bag of objects alone. Consistent with our previous observation, $w$ and $h$ (bounding box width and height) seems to be the most informative feature combination -- it performs well even without  positional information. Area ($a$) is less informative than the combination of $w$ and $h$, possibly because it compresses width-height ratio information despite discarding noise from background regions. Positional information ($x$, $y$) does not seem to be as informative, %
consistent with observations from previous work~\cite{WangGaizauskas:2016}.

\begin{table}[t]
\begin{center}
{\small
\begin{tabular}{l c c}
\toprule
Feature set & Fixed & Tuned \\
\midrule
Bag of objects & 0.807 & 0.834 \\ 
\midrule
($x, y, w, h, a$) & 0.870 & 0.915\\
($x, y, w, h$) & 0.859 & 0.898\\
($x, y, a$) & 0.850 & 0.900\\
($w, h$) & 0.870 & 0.920\\ 
($a$) & 0.869 & 0.857\\
($x, y$) & 0.810 & 0.863\\
\midrule
\multicolumn{2}{l}{LSTM \newcite{YinOrdonez:2017}$^\dagger$} & 0.922\\
\bottomrule
\end{tabular}
}
\end{center}
\caption{CIDEr scores for image captioning using representations encoding spatial information of instances derived from ground truth annotations, with either fixed hyperparameters (Section~\ref{sec:model}) or with hyperparameter tuning. $\dagger$~Results taken from \cite{YinOrdonez:2017}.}
\label{tbl:bboo-results}
\end{table}

The last column in Table~\ref{tbl:bboo-results} shows the CIDEr scores when  training the models by performing hyperparameter tuning during training. We note that the results with our simpler image representation are comparable to the ones reported in \newcite{YinOrdonez:2017}, which use more complex models to encode similar image information. Interestingly, we observe that positional information ($x$, $y$) work better than before tuning in this case.

Example outputs from the models in Sections~\ref{sec:boo} and \ref{sec:bboo} can be found in Figure~\ref{fig:examples}.

\begin{figure*}[!t]
  \begin{center}
    \includegraphics[width=0.90\linewidth]{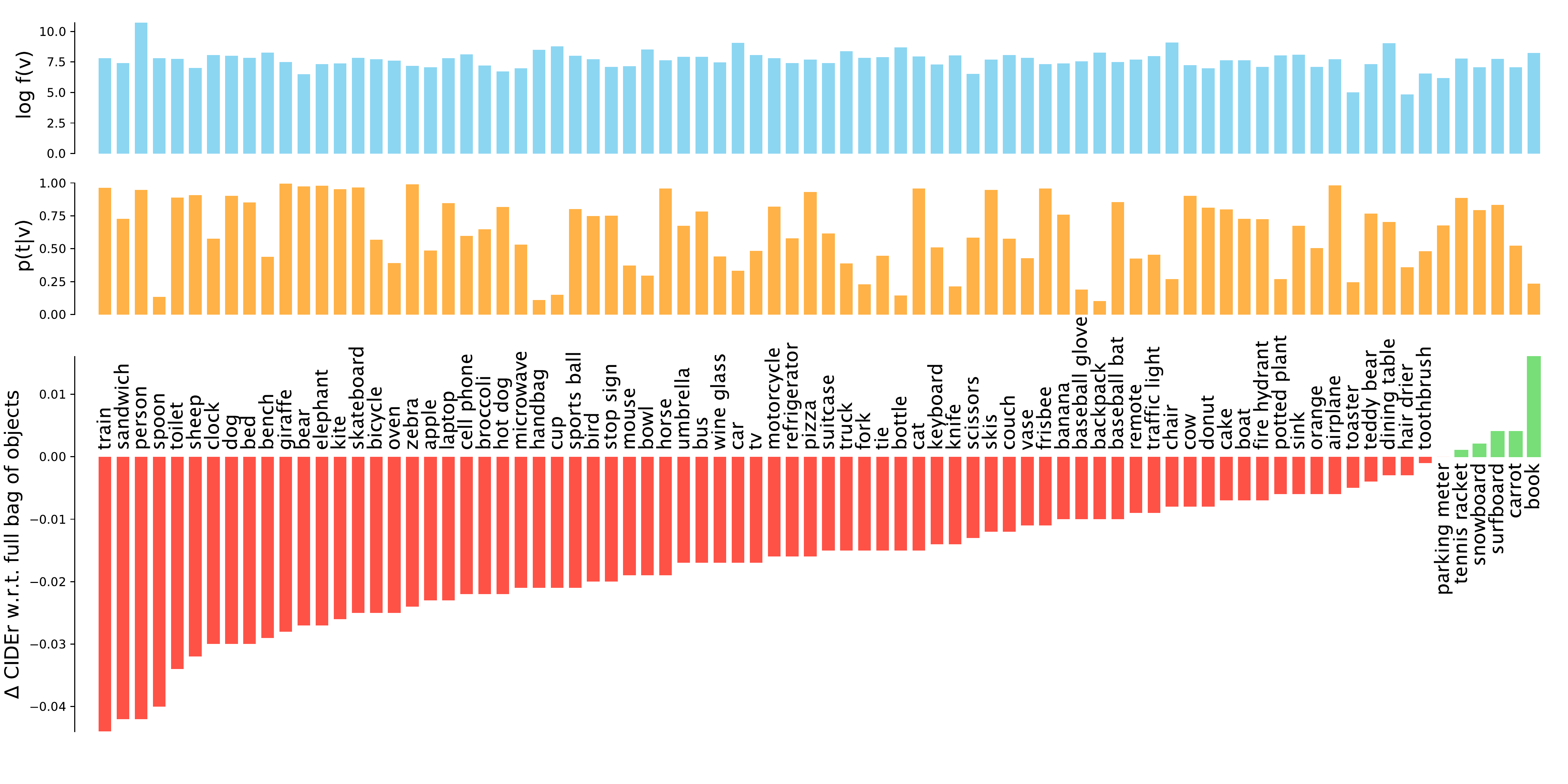}
    \caption{Difference in CIDEr scores wh/en removing each category from the bag of objects representation (79 dimensions), compared to using the full 80D vector (bottom plot). See main text for details.%
    }
    \label{fig:barplot_removeone}
  \end{center}
\end{figure*}

\section{Importance of different categories}
\label{sec:categories}

In the previous sections, we explore IC based on explicit detections for $80$ object categories. %
However, not all categories are made equal. Some categories could impact IC more than others~\cite{BergEtAl:2012}. In this section we investigate \textbf{\emph{which} categories are more important} for IC on the COCO dataset. Our \textbf{category ablation} experiment involves removing \emph{one} category from the $80$-dimensional bag of objects (ground truth frequency) representation at a time, resulting in $80$ sets of $79$D vectors without each ablated category. %

We postulate that salient categories %
should lead to larger performance degradation than others. %
However, what makes a category `salient' in general (\textit{dog} vs. \textit{cup})? We hypothesize that it could be due to (i) how frequently it is depicted across images; (ii) how frequently it is mentioned in the captions when depicted in the image. To quantify these hypotheses, we \textbf{compute the rank correlation} between changes in CIDEr from removing the category and each of the statistic below:

\begin{itemize}
\item $f(v_c) = \sum_{i}^{N} 1(c \in C_i)$: frequency of the ablated category $c$ being annotated among $N$ images in the training set, where $C_i$ is the set of all categories annotated in image $i$, and $1(x)$ is the indicator function.
\item $p(t_c|v_c) \approx \frac{f(t_c,v_c)}{f(v_c)}$: proportion of ablated category being mentioned in any of the reference captions given that it is annotated in the image in the training set.  
\end{itemize}

For determining whether a depicted category is mentioned in the caption, the matching method described in \newcite{RamisaEtAl:2015} is used to increase recall by matching category labels with (i) the term themselves; (ii) the head noun for multiword expressions; (iii) WordNet synonyms and hyponyms. We treat these statistics as an approximation because of the potential noise from the matching process, although it is clean enough for our purposes.

We have also tried computing the correlation with $f(t_c)$ (frequency of the %
category being mentioned regardless of whether or not it is depicted). However, we found the word matching process too noisy as it is not constrained or grounded on the image (e.g. ``hot dog'' is matched to the \emph{dog} category). Thus, we do not report the results for this.

\subsection{Experiments}

Figure~\ref{fig:barplot_removeone} shows the result of the category ablation experiment. Categories like \textit{train}, \textit{sandwich}, \textit{person} and \textit{spoon} led to the largest drop in CIDEr scores. On the other end, categories like %
\textit{surfboard}, \textit{carrot} and \textit{book} can be removed without negatively affecting the overall %
score. %

By comparing the CIDEr score changes against the frequency counts of object annotations in the training set (top row), there does not seem to be a clear correlation between depiction frequency and CIDEr. 
Categories like \textit{bear} %
are infrequent but led to a large drop in score; likewise, \textit{chair} and \textit{dining table} are frequent but do not affect the results as negatively.
In contrast, the frequency of a category being mentioned given that it is depicted is a better predictor for the changes in CIDEr scores in general (middle row). Animate objects seem to be important to IC and are often mentioned in captions \cite{BergEtAl:2012}. Interestingly, removing \textit{spoon} greatly affects the results even though it is not frequent in captions. %

Table~\ref{tbl:correlation} presents the rank correlation (Spearman's $\rho$ and Kendall's $\tau$, two-tailed test) between changes in CIDEr %
and the two heuristics. While both heuristics are positively correlated with the changes in CIDEr, we can conclude that the frequency of being mentioned (given that it is depicted) is better correlated with the score changes than the frequency of depiction. Of course, the categories are not mutually exclusive and object co-occurrence may also play a role. %
However, we leave this analysis for future work. %

\begin{table}
  \begin{center}
    { \small
    \begin{tabular}{l c c}
    \toprule
    & \multicolumn{2}{c}{Coefficient value ($p$-value)} \\
    \noalign{\smallskip}
    \cline{2-3}
    \noalign{\smallskip}
    & $f(v_c)$ & $p(t_c|v_c)$ \\
    \midrule
    Spearman's $\rho$ & 0.137 (0.226) & 0.227 (0.043) \\
    Kendall's $\tau$ & 0.093 (0.223) & 0.153 (0.047) \\
    \bottomrule
    \end{tabular}
    }
    \caption{Correlation between changes in CIDEr score from category ablation and the frequency of depiction of the category ($f(v_c)$) against the probably of it being mentioned in the caption given depiction ($(p(t_c|v_c)$).}
    \label{tbl:correlation}
  \end{center}
\end{table}

Figure~\ref{fig:examples} shows an example when the category \emph{person} is removed from the feature vector. Here, the model does not generate any text related to \emph{person}, as the training set contains images of clocks without people in it. %

\section{Conclusions}%
\label{sec:conclusion}

In this paper we investigated end-to-end image captioning by using highly interpretable representations derived from %
explicit object detections. We provided an in-depth analysis on the efficacy of a variety of cues derived from object detections for IC. We found that frequency counts, object size and position are informative and complementary. We also %
found that some categories have a bigger impact on IC than others. Our analysis showed that end-to-end IC systems are image matching systems that project image representations into a learned space and allow the LSTM to generate captions for images in that projected space. 

Future work includes (i) investigating how object category information can be better used or expanded to improve IC; (ii) analyzing end-to-end IC systems by using interpretable representations that rely on other explicit detectors (e.g. actions, scenes, attributes). The use of such explicit information about object instances could help improve our understanding of image captioning.

\section*{Acknowledgments}
This work is supported by the MultiMT project (H2020 ERC Starting Grant No. 678017). The authors also thank the anonymous reviewers for their valuable feedback on an earlier draft of the paper.

\bibliography{naacl2018}
\bibliographystyle{acl_natbib}

\appendix

\section{Hyperparmeter Settings}
\label{appendix:hyperparameter}

The hyperparameter settings for our model are as follows: 
\begin{itemize}
\item {\textbf{LSTM layers}}: $2$-Layer LSTM 
\item {\textbf{Word Embedding Dimensionality}}: 128
\item {\textbf{Hidden Layer Dimensionality}}: 256
\item {\textbf{Maximum Epochs}}: 50
\item {\textbf{Batch Size}}: [50, 100]
\item {\textbf{LSTM dropout settings}}: [0.2, 0.7]
\item {\textbf{Vocabulary threshold}}: 2
\item {\textbf{Learning Rate}}: [1e-4, 4e-4]
\item {\textbf{Optimiser}}: Adam
\end{itemize}

For items in a range of values, we used grid search to tune the hyperparmeters. %

\section{Full Experimental Results}
\label{appendix:fullresults}
Tables~\ref{tbl:boo-fullresults} and \ref{tbl:bboo-fullresults} show the results of several of our experiments with the most common metrics used in image captioning: BLEU, Meteor, ROUGE$_L$, CIDEr and SPICE. 

Figure~\ref{fig:object_distribution_hires} gives a high resolution version of Figure~\ref{fig:object_distribution}, showing the similarity between train and test distributions in terms of object categories.

\begin{table*}
\begin{center}
\begin{tabular}{lcccccccc}
\toprule
Representation & \textbf{B1} & \textbf{B2} & \textbf{B3} & \textbf{B4} & \textbf{M} & \textbf{R} & \textbf{C} & \textbf{S} \\
\midrule
ResNet-152 POOL5 & 0.664 & 0.480 & 0.335 & 0.233 & 0.220 & 0.486 & 0.749 & 0.150\\
\midrule
Frequency & 0.668 & 0.481 & 0.334 & 0.231 & 0.223 & 0.486 & 0.807 & 0.155\\
Normalized & 0.656 & 0.468 & 0.324 & 0.226 & 0.218 & 0.477 & 0.762 & 0.148\\
Binarized & 0.652 & 0.465 & 0.317 & 0.217 & 0.218 & 0.473 & 0.751 & 0.146\\
\midrule
Object min distance & 0.650 & 0.460 & 0.316 & 0.219 & 0.218 & 0.474 & 0.759 & 0.147\\
Object max size & 0.661 & 0.476 & 0.332 & 0.232 & 0.224 & 0.483 & 0.793 & 0.151\\
\midrule
Obj max size + Obj min distance & 0.670 & 0.482 & 0.333 & 0.231 & 0.225 & 0.485 & 0.799 & 0.153\\
Frequency + Obj min distance & 0.675 & 0.491 & 0.345 & 0.239 & 0.229 & 0.495 & 0.836 & 0.160\\
Frequency + Obj max size & 0.684 & 0.496 & 0.349 & 0.244 & 0.228 & 0.495 & 0.830 & 0.159\\
All three features & 0.683 & 0.501 & 0.355 & 0.250 & 0.229 & 0.498 & 0.849 & 0.162\\
\bottomrule
\end{tabular}
\end{center}
\caption{Full results for image captioning using ground truth bag of objects variants as visual representations, for metrics \textbf{B}LEU, \textbf{M}eteor, \textbf{R}OUGE$_L$, \textbf{C}IDEr and \textbf{S}PICE.}
\label{tbl:boo-fullresults}
\end{table*}

\begin{table*}
\begin{center}
\begin{tabular}{lcccccccc}
\toprule
Representation & \textbf{B1} & \textbf{B2} & \textbf{B3} & \textbf{B4} & \textbf{M} & \textbf{R} & \textbf{C} & \textbf{S} \\
\midrule
Bag of objects & 0.668 & 0.481 & 0.334 & 0.231 & 0.223 & 0.486 & 0.807 & 0.155\\
\midrule
($x, y, w, h, a$) & 0.683 & 0.503 & 0.359 & 0.255 & 0.233 & 0.503 & 0.870 & 0.163\\
($x, y, w, h$) & 0.687 & 0.503 & 0.355 & 0.251 & 0.233 & 0.501 & 0.859 & 0.166\\
($x, y, a$) & 0.683 & 0.502 & 0.356 & 0.250 & 0.232 & 0.501 & 0.850 & 0.164\\
($w, h$) & 0.693 & 0.511 & 0.364 & 0.256 & 0.233 & 0.505 & 0.870 & 0.165\\
($a$) & 0.684 & 0.503 & 0.358 & 0.254 & 0.232 & 0.501 & 0.869 & 0.162\\
($x, y$) & 0.675 & 0.488 & 0.341 & 0.237 & 0.224 & 0.491 & 0.810 & 0.155\\
\bottomrule
\end{tabular}
\end{center}
\caption{Full results for image captioning, using representations encoding spatial information of instances derived from ground truth annotations with fixed hyperparameters, for metrics \textbf{B}LEU, \textbf{M}eteor, \textbf{R}OUGE$_L$, \textbf{C}IDEr and \textbf{S}PICE.}
\label{tbl:bboo-fullresults}
\end{table*}

\begin{figure*}
  \begin{center}
  \includegraphics[width=1.0\linewidth]{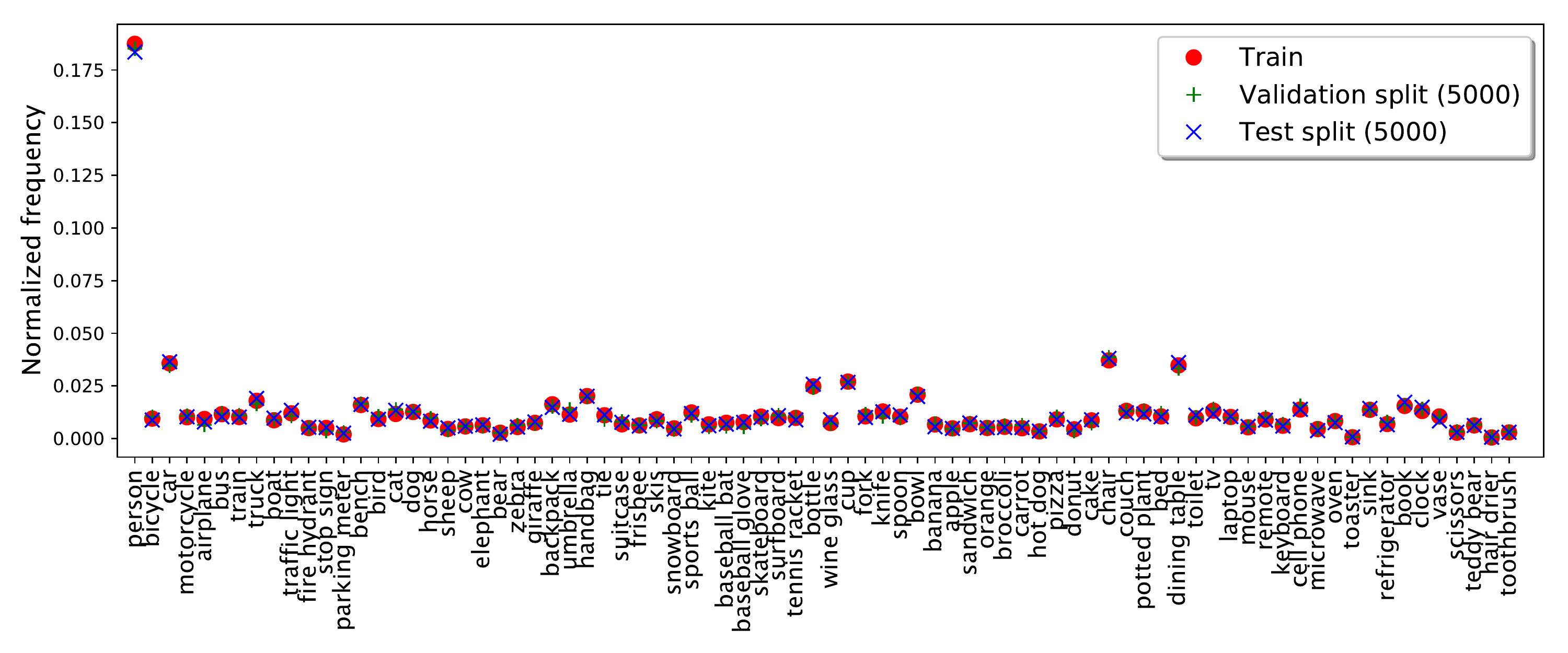}
  \end{center}
  \vspace{-18pt}
  \caption{Object category distributions for COCO train, validation and test splits:  normalized document frequency of each category. The distribution between the training and test sets are almost identical.}
  \label{fig:object_distribution_hires}
\end{figure*}

\section{Example captions for different models}
\label{appendix:example-output}

Figure~\ref{fig:rep-analysis} shows example images from COCO and the output captions from different models. We compare the outputs of selected models from Sections \ref{sec:boo} and \ref{sec:bboo}, and a model where the \emph{person} category is removed from the input vector (Section~\ref{sec:categories}).

\section{Example nearest neighbours for test images}
\label{appendix:knn-moreexample}

Figure~\ref{fig:knn-moreexample} shows the five nearest neighbours in the training set of each non-replica example from the test set (where the exact ground truth frequency representation does not occur in training). See Section~\ref{sec:analysis} for a more detailed description of the experiment. 

\begin{figure*}
{\scriptsize
\begin{minipage}{0.2\textwidth}
\includegraphics[width=0.9\textwidth]{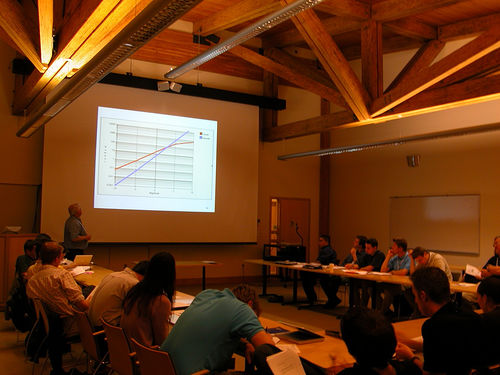}
Image ID: 165225
\end{minipage}
\begin{minipage}{0.8\textwidth}
\begin{tabular}{p{0.25\linewidth}p{0.65\linewidth}}
\toprule
\textbf{Objects in the image} & \emph{person}, \emph{tie}, \emph{tv}, \emph{laptop}, \emph{chair}\\
\midrule
\textbf{Representation} & \textbf{Caption}\\
\midrule
Frequency & a group of people sitting around a table with a laptop . \\
Object min distance & a man sitting at a desk with a laptop computer . \\
Object max size &  a man standing in front of a tv in a living room .\\
All three features &  a group of people sitting at a table with a laptop . \\
$(x,y)$ &  a group of people standing around a table with a microphone . \\
$(w,h)$ & a group of people sitting around a table with a laptop .\\
$(a)$ & a group of people standing in a living room . \\
$(x,y,w,h,a)$ & a group of people sitting at a table with laptops .\\
\cdashlinelr{1-2}
CNN (ResNet-$152$) & a group of people sitting around the table .\\
\textit{person} removed & a man standing in front of a laptop computer .\\
\bottomrule
\end{tabular}
\end{minipage}
}

{\scriptsize
\vspace{0.6cm}
\begin{minipage}{0.2\textwidth}
\includegraphics[width=0.9\textwidth]{COCO_val2014_000000196715.jpg}
Image ID: 196715
\end{minipage}
\begin{minipage}{0.8\textwidth}
\begin{tabular}{p{0.25\linewidth}p{0.65\linewidth}}
\toprule
\textbf{Objects in the image}  & \emph{person}, \emph{car}, \emph{truck}, \emph{surfboard}\\
\midrule
\textbf{Representation} & \textbf{Caption}\\
\midrule
Frequency &  a man riding a surfboard on top of a wave . \\
Object min distance &  a man riding a surfboard on top of a wave .\\
Object max size &  a red and white truck driving down a street . \\
All three features & a red and white truck driving down a street . \\
$(x,y)$ & a large white car parked in a parking lot .  \\
$(w,h)$ &  a large white truck with a surfboard on the back of it .\\
$(a)$ &  a man riding a horse on a beach next to a dog . \\
$(x,y,w,h,a)$ & a car is parked on the side of a street with a car in the background .\\
\cdashlinelr{1-2}
CNN (ResNet-$152$) & a boat is parked on the side of the road .\\
\textit{person} removed & a man is standing on a surfboard in the water .\\
\bottomrule
\end{tabular}
\end{minipage}
}

{\scriptsize
\vspace{0.6cm}
\begin{minipage}{0.2\textwidth}
\includegraphics[width=0.9\textwidth]{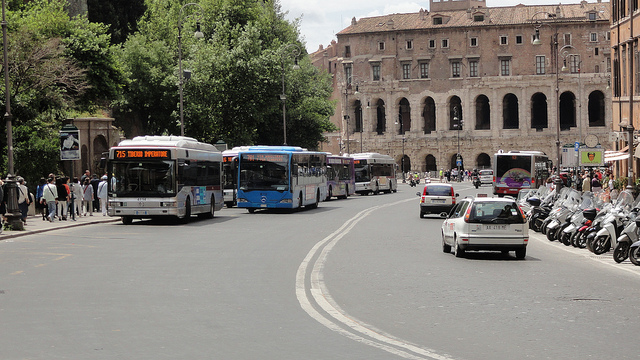}
Image ID: 491793
\end{minipage}
\begin{minipage}{0.8\textwidth}
\begin{tabular}{p{0.25\linewidth}p{0.65\linewidth}}
\toprule
\textbf{Objects in the image}  & \emph{person}, \emph{car}, \emph{motorcycle}, \emph{bus}, \emph{handbag} \\
\midrule
\textbf{Representation} & \textbf{Caption}\\
\midrule
Frequency & a group of people riding motorcycles down a street . \\
Object min distance & a city bus driving down a street next to a traffic light . \\
Object max size & a bus is driving down the street in the city . \\
All three features &  a group of people standing around a parking lot . \\
$(x,y)$ &  a group of people standing around motorcycles in a parking lot . \\
$(w,h)$ & a group of people standing around a bus stop .\\
$(a)$ &  a group of people riding bikes down a street .\\
$(x,y,w,h,a)$ & a group of motorcycles parked next to each other on a street .\\ 
\cdashlinelr{1-2}
CNN (ResNet-$152$) & a bus is driving down a street with a lot of people .\\
\textit{person} removed & a group of people riding motorcycles on a city street .\\
\bottomrule
\end{tabular}
\end{minipage}
}

{\scriptsize
\vspace{0.6cm}
\begin{minipage}{0.2\textwidth}
\includegraphics[width=0.9\textwidth]{COCO_val2014_000000378657.jpg}
Image ID: 378657
\end{minipage}
\begin{minipage}{0.80\textwidth}
\begin{tabular}{p{0.25\linewidth}p{0.65\linewidth}}
\toprule
\textbf{Objects in the image}  & \emph{person}, \emph{clock} \\
\midrule
\textbf{Representation} & \textbf{Caption}\\
\midrule
Frequency & a large clock tower with a large clock on it . \\
Object min distance & a clock tower with a large clock on it 's face . \\
Object max size & a man standing in front of a clock tower . \\
All three features &  a clock tower with people standing in the middle of the water . \\
$(x,y)$ &  a large clock tower with a clock on the front . \\
$(w,h)$ & a clock on a pole in front of a building\\
$(a)$ &  a large clock tower with people walking around it\\
$(x,y,w,h,a)$ & a group of people standing around a clock tower .\\ 
\cdashlinelr{1-2}
CNN (ResNet-$152$) & a large building with a clock tower in the middle of it .\\
\textit{person} removed & a clock tower with a weather vane on top of it .\\
\bottomrule
\end{tabular}
\end{minipage}
}

\caption{Examples of descriptions where models differ. The models with explicit object detection and additional spatial information ($(x,y,w,h,a)$) is more precise in most cases (even though still incorrect in the second example). In the first example, aggregating multiple instances for size and distance cues clearly removes the information about the group of people in the image. 
The output of a standard CNN (ResNet-152 POOL5) is also shown, as well as that of the model where the most salient category -- \textit{person} -- is removed.
}
\label{fig:rep-analysis}
\end{figure*}

\begin{figure*}
\begin{center}
{ \scriptsize
\begin{tabular}{l c l}
\multicolumn{3}{c}{\textbf{Test Image ID: 242946}}\\
\toprule
\multirow{4}{*}{\begin{sideways}test\end{sideways}} & \multirow{4}{*}{\includegraphics[height=1.0cm]{COCO_val2014_000000242946.jpg}} & \\
& & \textit{person~(5), cup~(8), spoon~(1), bowl~(8), carrot~(10), chair~(6), dining table~(3)} \\
& & \vspace{-2pt} $\Rightarrow$ a group of people sitting around a table with food .\\
 & & \\
\midrule
\multirow{4}{*}{1} & \multirow{4}{*}{\includegraphics[height=1.0cm]{COCO_train2014_000000280829.jpg}} & \\
& & \textit{person~(4), cup~(4), spoon~(1), bowl~(5), chair~(6), dining table~(4)} \\
& & \vspace{-2pt} $\Rightarrow$ a woman sitting at a table with a plate of food .\\
 & & \\
\multirow{4}{*}{2} & \multirow{4}{*}{\includegraphics[height=1.0cm]{COCO_train2014_000000157186.jpg}} & \\
& & \textit{person~(9), bottle~(1), cup~(6), bowl~(4), broccoli~(2), chair~(5), dining table~(3)} \\
& & \vspace{-2pt} $\Rightarrow$ a group of people sitting at a table eating food .\\
 & & \\
\multirow{4}{*}{3} & \multirow{4}{*}{\includegraphics[height=1.0cm]{COCO_train2014_000000506555.jpg}} & \\
& & \textit{person~(11), cup~(2), bowl~(4), carrot~(6), cake~(1), chair~(4), dining table~(1)} \\
& & \vspace{-2pt} $\Rightarrow$ a group of people sitting around a table with a cake .\\
 & & \\
\multirow{4}{*}{4} & \multirow{4}{*}{\includegraphics[height=1.0cm]{COCO_train2014_000000093992.jpg}} & \\
& & \textit{cup~(1), spoon~(1), bowl~(9), carrot~(10), chair~(3), potted plant~(1), dining table~(1), vase~(1)} \\
& & \vspace{-2pt} $\Rightarrow$ a table with a variety of food on it .\\
 & & \\
\multirow{4}{*}{5} & \multirow{4}{*}{\includegraphics[height=1.0cm]{COCO_train2014_000000406753.jpg}} & \\
& & \textit{cup~(4), bowl~(7), carrot~(6), dining table~(1)} \\
& & \vspace{-2pt} $\Rightarrow$ a table with bowls of food and vegetables .\\
 & & \\
\end{tabular}

\begin{tabular}{l c l}
\noalign{\bigskip}
\noalign{\bigskip}
\multicolumn{3}{c}{\textbf{Test Image ID: 378962}}\\
\toprule
\multirow{4}{*}{\begin{sideways}test\end{sideways}} & \multirow{4}{*}{\includegraphics[height=1.0cm]{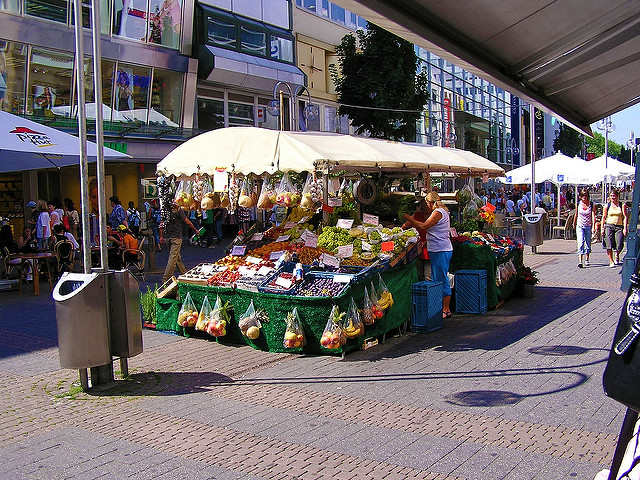}} & \\
& & \textit{person~(14), backpack~(3), umbrella~(4), handbag~(1), banana~(4), apple~(6), orange~(10), chair~(7), dining table~(2)} \\
& & \vspace{-2pt} $\Rightarrow$ a group of people standing around a fruit stand .\\
 & & \\
\midrule
\multirow{4}{*}{1} & \multirow{4}{*}{\includegraphics[height=1.0cm]{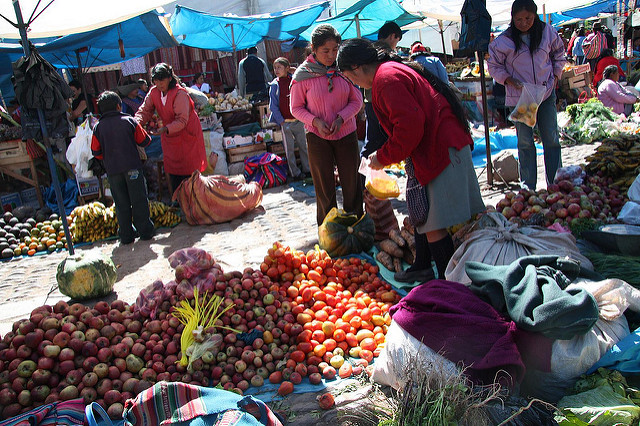}} & \\
& & \textit{person~(14), backpack~(1), banana~(5), apple~(5), orange~(13)} \\
& & \vspace{-2pt} $\Rightarrow$ a group of people standing around a fruit stand .\\
 & & \\
\multirow{4}{*}{2} & \multirow{4}{*}{\includegraphics[height=1.0cm]{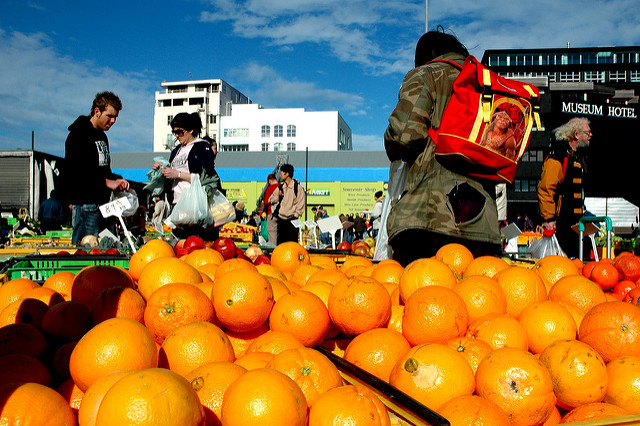}} & \\
& & \textit{person~(13), truck~(1), backpack~(1), apple~(2), orange~(10)} \\
& & \vspace{-2pt} $\Rightarrow$ a group of people standing around a fruit stand .\\
 & & \\
\multirow{4}{*}{3} & \multirow{4}{*}{\includegraphics[height=1.0cm]{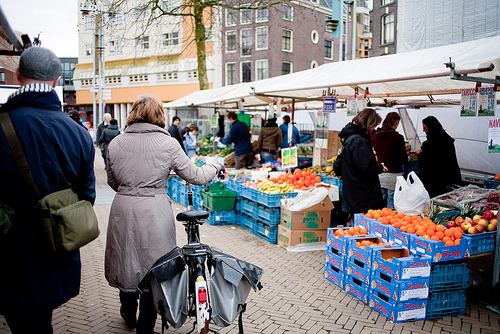}} & \\
& & \textit{person~(12), bicycle~(1), handbag~(3), banana~(2), apple~(2), orange~(7)} \\
& & \vspace{-2pt} $\Rightarrow$ a group of people standing around a fruit stand .\\
 & & \\
\multirow{4}{*}{4} & \multirow{4}{*}{\includegraphics[height=1.0cm]{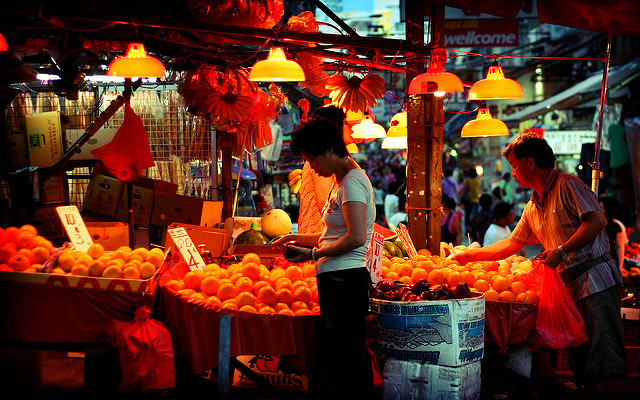}} & \\
& & \textit{person~(11), banana~(4), apple~(1), orange~(14)} \\
& & \vspace{-2pt} $\Rightarrow$ a man standing next to a fruit stand with bananas .\\
 & & \\
\multirow{4}{*}{5} & \multirow{4}{*}{\includegraphics[height=1.0cm]{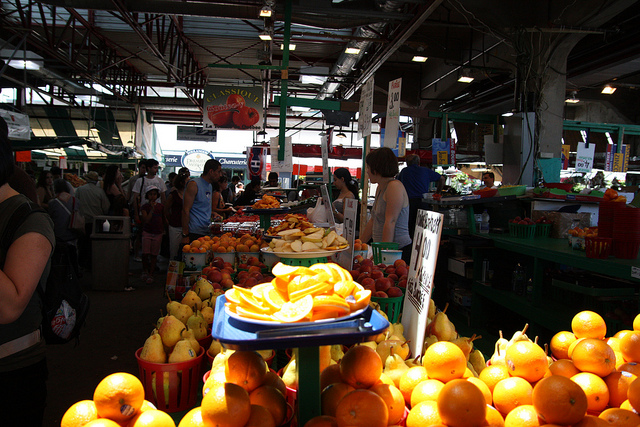}} & \\
& & \textit{person~(14), backpack~(1), handbag~(3), apple~(7), orange~(14)} \\
& & \vspace{-2pt} $\Rightarrow$ a group of people standing around a fruit stand .\\
 & & \\
\end{tabular}

\begin{tabular}{l c l}
\noalign{\bigskip}
\noalign{\bigskip}
\multicolumn{3}{c}{\textbf{Test Image ID: 223648}}\\
\toprule
\multirow{4}{*}{\begin{sideways}test\end{sideways}} & \multirow{4}{*}{\includegraphics[height=1.0cm]{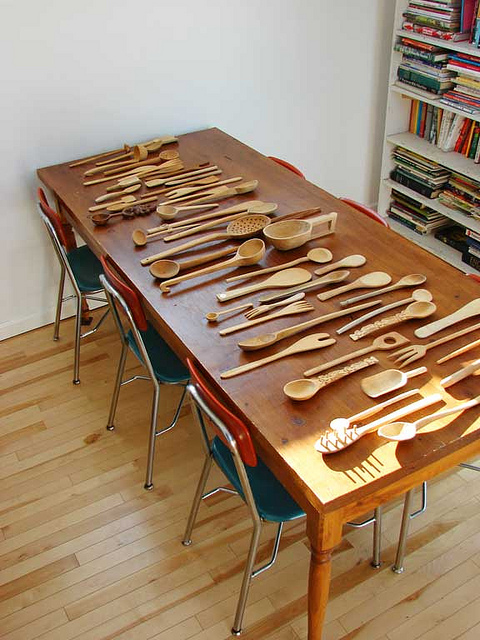}} & \\
& & \textit{fork~(3), spoon~(14), chair~(6), dining table~(1), book~(14)} \\
& & \vspace{-2pt} $\Rightarrow$ a dining room with a table and chairs .\\
 & & \\
\midrule
\multirow{4}{*}{1} & \multirow{4}{*}{\includegraphics[height=1.0cm]{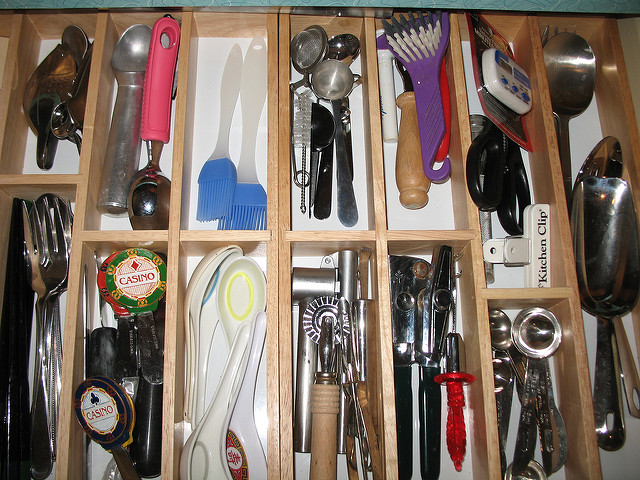}} & \\
& & \textit{fork~(1), knife~(1), spoon~(13), scissors~(1)} \\
& & \vspace{-2pt} $\Rightarrow$ a drawer of a variety of different types of food .\\
 & & \\
\multirow{4}{*}{2} & \multirow{4}{*}{\includegraphics[height=1.0cm]{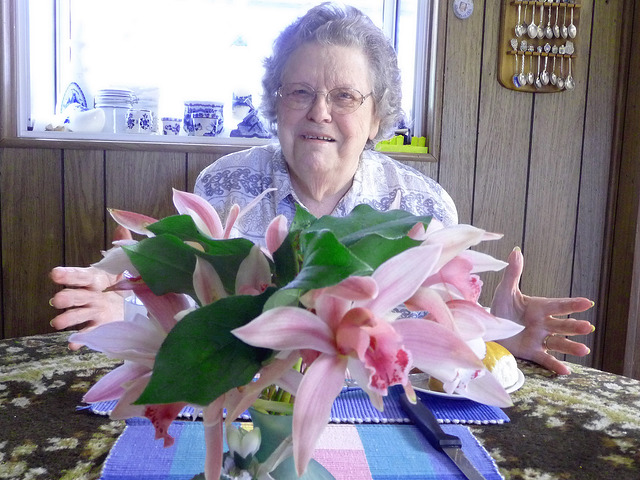}} & \\
& & \textit{person~(1), cup~(3), knife~(1), spoon~(14), bowl~(1), potted plant~(1), dining table~(1), vase~(1)} \\
& & \vspace{-2pt} $\Rightarrow$ a woman is sitting at a table with a glass of wine .\\
 & & \\
\multirow{4}{*}{3} & \multirow{4}{*}{\includegraphics[height=1.0cm]{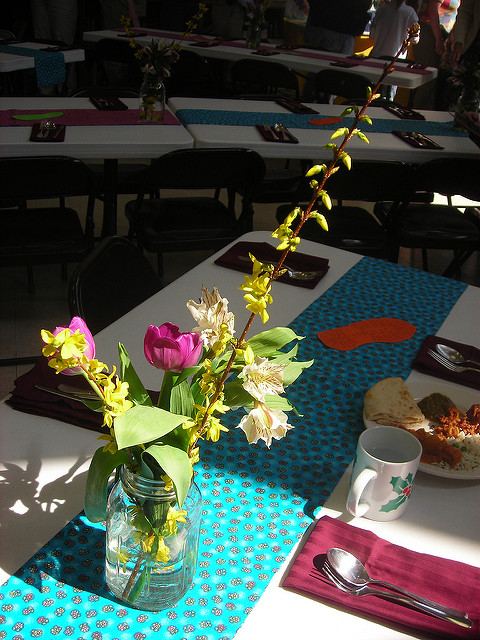}} & \\
& & \textit{person~(1), cup~(1), fork~(3), spoon~(6), chair~(9), dining table~(5), vase~(1)} \\
& & \vspace{-2pt} $\Rightarrow$ a woman sitting at a table with a plate of food .\\
 & & \\
\multirow{4}{*}{4} & \multirow{4}{*}{\includegraphics[height=1.0cm]{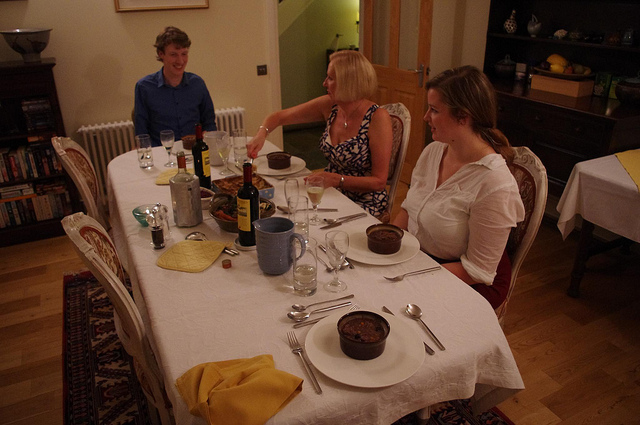}} & \\
& & \textit{person~(3), bottle~(3), wine glass~(5), cup~(4), fork~(3), knife~(2), spoon~(11), bowl~(5), chair~(4), dining table~(2), book~(
2)} \\
& & \vspace{-2pt} $\Rightarrow$ a group of people sitting around a table with food .\\
 & & \\
\multirow{4}{*}{5} & \multirow{4}{*}{\includegraphics[height=1.0cm]{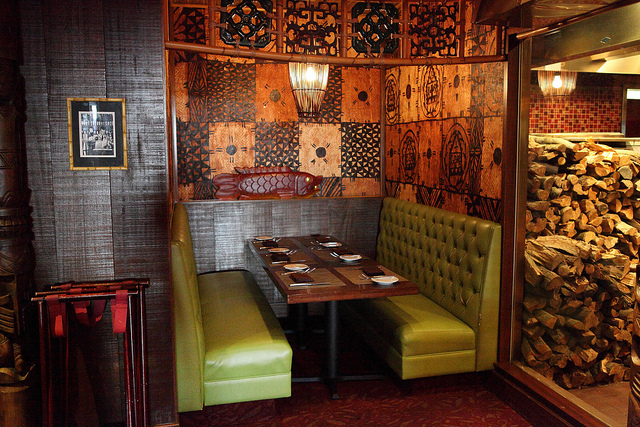}} & \\
& & \textit{fork~(4), knife~(4), spoon~(7), chair~(1), couch~(2), dining table~(1)} \\
& & \vspace{-2pt} $\Rightarrow$ a table with a table and chairs and a table .\\
 & & \\
\end{tabular}

}
\end{center}
\caption{Five nearest neighbours from the training set in the projected space, for several example test images' (top row of each table) original bag of objects representation that does not have an exact match in the training. For each image, we show the ground truth categories (and frequencies in parenthesis) and the generated caption.}
\label{fig:knn-moreexample}
\end{figure*} 
\end{document}